\documentclass[runningheads]{llncs}
\usepackage{graphicx}
\usepackage{hyperref}

\usepackage{tikz}
\usepackage{comment}
\usepackage{amsmath,amssymb} 
\usepackage{color}
\usepackage{rotating}
\usepackage{multirow, makecell}
\usepackage{subcaption}

\begin{document}
\pagestyle{headings}
\mainmatter
\def\ECCVSubNumber{7}  

\title{Emotion Understanding in Videos Through Body, Context, and Visual-Semantic Embedding Loss} 


\titlerunning{Emotion Understanding in Videos}

\author{Panagiotis Paraskevas Filntisis\inst{1} \and
Niki Efthymiou\inst{1} \and
Gerasimos Potamianos\inst{2} \and
Petros Maragos\inst{1}
}
\authorrunning{P.P. Filntisis et al.}
%
\institute{School of E.C.E., NTUA, Greece \and  E.C.E. Department, UTH, Greece \\
\email{\{filby,nefthymiou\}@central.ntua.gr, gpotam@ieee.org, maragos@cs.ntua.gr}}
\maketitle

\begin{abstract}
We present our winning submission to the First International Workshop on Bodily Expressed Emotion Understanding (BEEU) challenge. Based on recent literature on the effect of context/environment on emotion, as well as visual representations with semantic meaning using word embeddings, we extend the framework of Temporal Segment Network to accommodate these. Our method is verified on the validation set of the Body Language Dataset (BoLD) and achieves 0.26235 Emotion Recognition Score on the test set, surpassing the previous best result of 0.2530.
\keywords{emotion, body, context, visual-semantic, BEEU challenge}
\end{abstract}

\section{Introduction}
Automatic human affect recognition from visual cues is an important area of computer vision that has attracted increased interest over the last two decades, due to its many applications. Indeed, social robotics \cite{cavallo2018emotion}, psychiatric care \cite{gaudelus2016improving}, and edutainment \cite{filntisis2019fusing} are all areas that can benefit from automatic recognition of emotion.

Most past approaches to the problem have focused on facial expressions in order to determine the emotional state of the person of interest \cite{du2014compound,lucey2010extended,mollahosseini2017affectnet}. This is reasonable due to the fact that facial expressions have been studied extensively in the psychology and emotion literature  \cite{ekman1997universal}. For example, the Facial Action Coding System (FACS) \cite{ekman1997face} identifies the units of facial movements, based on facial muscle groups. Combinations of the so-called action units (AUs) have also been linked with emotional states with extensions of the basic FACS such as EMFACS (Emotion FACS) \cite{friesen1983emfacs}. On the other hand, there is no similar established coding system for body expressions, although some have been proposed~\cite{dael2012body}.


Compared to facial expression based approaches, recent works have sought alternative modalities and streams of information to detect emotion; one is bodily expressions since many have highlighted the fact that the emotional state is conveyed through bodily expressions as well, and in certain emotions  it is the main modality \cite{de2009bodies,kleinsmith13,tracy2004show}, or can be used to correctly disambiguate the corresponding facial expression \cite{aviezer2012body}. Simultaneously, it is important to note that in cases and applications where the emotion needs to be identified, the human body is more frequently available than the face since the face can be occluded, hidden, or far in the distance. Another auxiliary stream of information besides the face and the body that can help in identifying emotions is the context and the surrounding environment of the person \cite{kosti_emotion_2017,mittal2020emoticon}. It is apparent that both the place, as well as objects and other humans can influence a person's emotions. 


We should also note that inherently emotion recognition is a multi-label problem - the subject might be feeling two or more emotions. This is true, especially when considering an extended set of emotions, as in \cite{luo_arbee_2020}. The emotions in extended sets do not have the same ``semantic" distance between them. For example, anger is more close to annoyance than to happiness. Considering that previous works have showed the superiority of methods that attempt to learn a joint embedding space that contains both word embeddings and visual representations \cite{dong2016word2visualvec,frome2013devise,ren2017multiple}, we believe that trying to attach a semantic meaning to the extracted visual feature is a natural way forward.

In this paper, based on the above, we describe the method of our team in the First International Workshop on Bodily Expressed Emotion Understanding (BEEU) challenge. Our method combines Temporal Segment Networks (TSNs) \cite{wang2016temporal} focusing on the body, using the context in each video as an additional stream, and also uses an extra visual-semantic embedding loss, based on GloVE (Global Vectors) \cite{pennington2014glove} word embedding representations. Our experiments in the validation set verify the better performance of our method compared to the traditional TSNs, while our emotion recognition score on the test set was 0.26235.




\section{Related Work}
While most past approaches in visual detection of affect have been focused on facial expressions \cite{de2009bodies}, recent approaches have started taking into account the body language \cite{kleinsmith13} of the person in question, as well as its surrounding context/environment.

In \cite{gunes2006bimodal}, Gunes and Piccardi introduced a bimodal architecture that takes into account both upper body and facial expressions, in order to detect affect in videos. In \cite{dael2012emotion}, Dael et al. analyzed and classified body emotional expressions using a body action and posture coding system which was proposed in \cite{dael2012body}. The 3D pose of children was also utilized in \cite{marinoiu20183d} by Marinoui et al. to detect emotions in continuous dimensions, while in \cite{filntisis2019fusing}, 2D pose was used and fused with facial expressions for child emotion recognition. Luo et al. \cite{luo_arbee_2020} introduced a large scale video dataset (BoLD) annotated with categorical and continuous emotions, which is the one used in the BEEU challenge.

Regarding the context modality,  Kosti et al. \cite{kosti_emotion_2017} introduced a large scale dataset for emotion recognition (EMOTIC) in different contexts (e.g., other people, places, or objects) and a convolutional neural network (CNN) based two-stream architecture that focused on the body and context of the subjects. The CAER video dataset for context-based emotion recognition was presented in \cite{lee2019context}, along with a two-stream architecture which employed adaptive-fusion to merge the two steams. In \cite{mittal2020emoticon}, Mittal et al. designed a deep architecture with several branches, focusing on different interpretations of the surrounding context (e.g., environment and interaction context) to significantly increase resulting predictions in the EMOTIC dataset.

Finally, some recent works have also focused on extracting visual representations from images that present the semantic relations found in embeddings built from words. The DeViSE embedding model \cite{frome2013devise} extracted semantically-meaningful visual representations by introducing a similarity loss between the feature vector extracted from a CNN and the word embedding from a skip-gram text model. Using a similar method, Wei et al. \cite{wei2020learning} built joint text and visual embeddings as emotion representation from web images, and in \cite{yeh_multilabel_2020}, Ye and Li built semantic embeddings for a multi-label classification problem.

\section{Dataset}

The dataset used in the challenge is the BoLD (Body Language Dataset) corpus \cite{luo_arbee_2020} consisting of 9,876 video clips of humans expressing emotion, primarily through body movements. Each clip can contain multiple characters, yielding a total of 13,239 annotations, split into a training, validation, and test set. The dataset has been annotated by crowdsourcing employing two widely accepted categorizations of emotion. The first one is the categorical annotation with a total of 26 labels first used in \cite{kosti_emotion_2017}, by collecting and processing an extensive affective vocabulary. The second annotation regards the continuous emotional dimensions of the VAD (Valence - Arousal - Dominance) Emotional State Model \cite{russell_evidence_1977}. The methods in the challenge are evaluated using the following Emotion Recognition Score (ERS):

\begin{equation}
    \mathrm{ERS}=\frac{1}{2}\left(\mathrm{m} R^{2}+\frac{1}{2}(\mathrm{mAP}+\mathrm{mRA})\right)
\end{equation}

\noindent where $\mathrm{m} R^{2}$ is the mean coefficient of determination ($R^2$) score for the three dimensional emotions (VAD), and $\mathrm{mAP}$ and $\mathrm{mRA}$ is the mean Average Precision and the mean area under receiver operating characteristic curve (ROC AUC) of the multilabel categorical predictions.

\begin{figure}[h]
\centering
\includegraphics[width=0.9\linewidth]{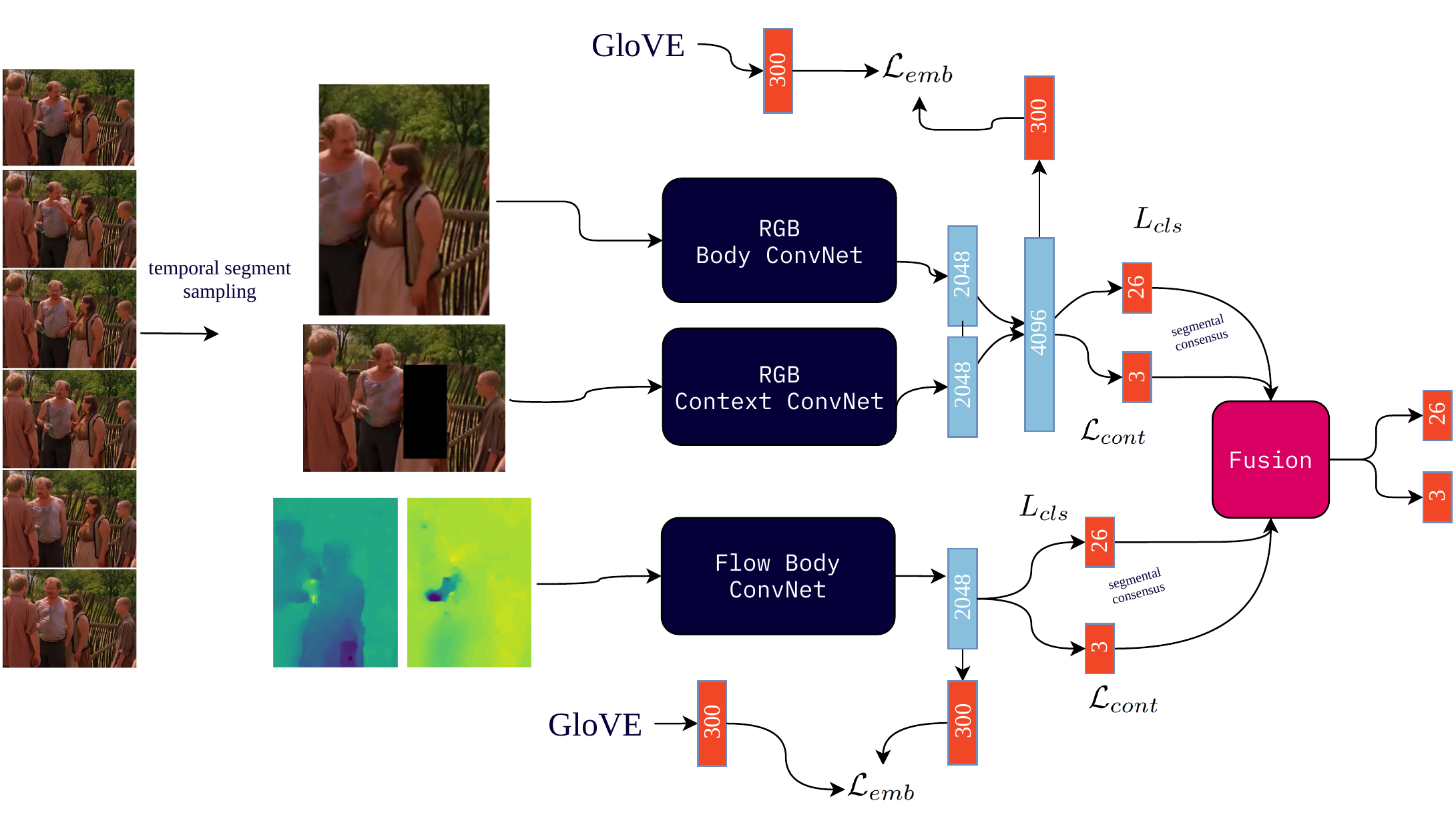} 
\caption{TSN with two RGB spatial streams (body and context) and one optical flow stream. The final results are obtained using average score fusion.}
\label{fig:arch}
\end{figure}

\section{Model Architecture}
Our model is based on the TSN architecture \cite{wang2016temporal}, which has been widely used in action recognition and can be seen in Fig. \ref{fig:arch}. During training, $K$ different segments are selected from the input video, and then $N$ consecutive frames are selected from each segment. This is done to deal with the fact that consecutive frames have usually redundant information. Traditionally, two different modalities are used, one is the spatial (RGB) modality and the second one is the optical flow. TSNs have already been shown to achieve good results for the BoLD dataset in its introductory paper~\cite{luo_arbee_2020}. 


In our approach, we modify the original version of TSNs mainly in two directions:

\paragraph{Context:}
We introduce one additional stream based on the context-environment surrounding the annotated human. For the RGB modality, we input the context in the network in the same way as in \cite{mittal2020emoticon}, by masking out the instance body (we set all pixels to 0). We call this stream RGB-c, and the body streams RGB-b and Flow-b. During training, the RGB-b and RGB-c streams are combined at the feature level (RGB-bc) and are trained jointly while the Flow-b TSN is trained independently.


\paragraph{Embedding Loss:}
Our second extension is the introduction of an embedding loss on the feature vector extracted by the Convolutional Neural Network (ConvNet). This is done to exploit the fact that some emotions are closer semantically to others. This is also revealed by examining the correlation matrix of the dataset labels in \cite{luo_arbee_2020}, where some labels occur more frequently in combination with others (e.g. Happiness and Pleasure, Annoyance and Anger, etc.). Due to this result, we try to attach a semantic meaning to the feature vector extracted by the backbone image network.

\begin{figure}[t!]
\centering
\includegraphics[width=0.9\linewidth]{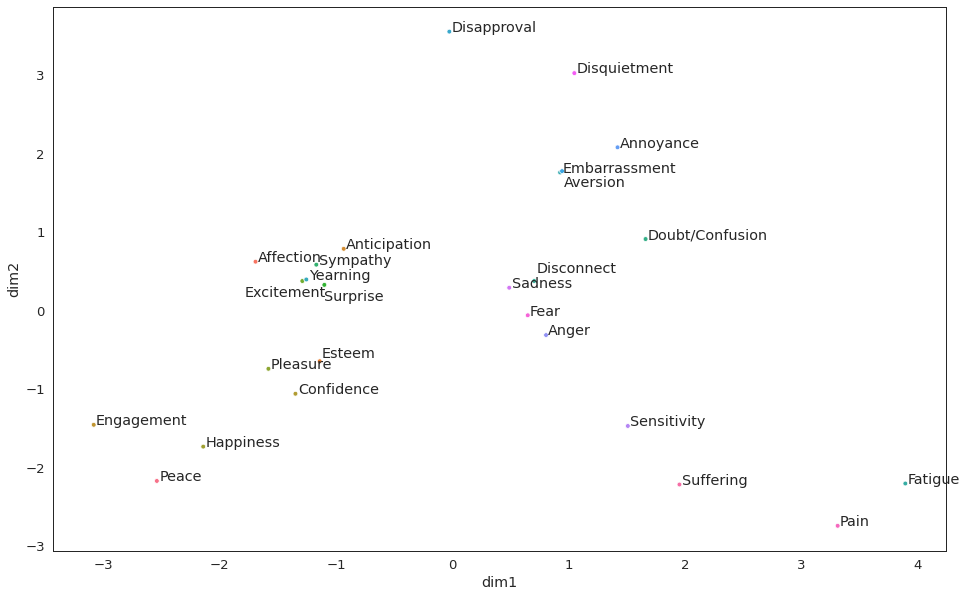}
\caption{PCA projection of the categorical emotions GloVE word embeddings.}
\label{fig:pca}
\end{figure}

To implement this, we first obtain for each one of the 26 categorical labels of BoLD their 300-dimensional GloVE word embedding \cite{pennington2014glove}. A PCA-projection of the 26 embeddings is shown in Fig.~\ref{fig:pca}, where it is apparent that the distances between embeddings are indicative of their ``semantic" distance. We then use a fully connected layer to map the feature extracted from the image to a 300-dimensional space and introduce the following mean-squared based loss:

\begin{equation}
    \mathcal{L}_{emb} = ||\boldsymbol{W} f_{v}(\boldsymbol{x}) - \frac{1}{|K|} \sum_{y \in K} f_{w}(\boldsymbol{y})||_{2}
    \label{eq:embedloss}
\end{equation}

\noindent where $f_{v}(\boldsymbol{x})$ is the feature vector extracted by applying the convNet on the image $\boldsymbol{x}$, $\boldsymbol{W}$ is a linear transformation from the space of the feature vector to the word embedding space, $f_{w}(\boldsymbol{y})$ is the word embedding of the label $y$, and $K$ is the set of all positive labels for the image $\boldsymbol{x}$. That is, we try to reduce the Euclidean distance between the projected image feature and the arithmetic mean of the GloVE embeddings of the positive labels for image/video.

\paragraph{Predictions:}
Finally, after extracting for each sampled image its feature vector, we use two fully connected layers, one to classify to the 26 different categorical labels, and one to regress over the 3 different categorical emotions. The two TSNs are trained using the following loss:

\begin{equation}
\mathcal{L} = \mathcal{L}_{cls_{1}} + \mathcal{L}_{cls_{2}} + \mathcal{L}_{cont} + \mathcal{L}_{emb}
\end{equation}


\noindent Specifically, since the dataset does not provide explicitly the multilabel targets, but the crowdsourced scores between 0 and 1, we include two different losses for the classification part: $\mathcal{L}_{cls_{1}}$ that is the binary cross-entropy between the predicted scores and the multilabel target (obtained after thresholding the multilabel scores at 0.5) and $\mathcal{L}_{cls_{2}}$ that is the mean squared error between the predicted scores and the multilabel scores. We empirically found that the inclusion of $\mathcal{L}_{cls_{2}}$ slightly boosted performance. For the regression part, $\mathcal{L}_{cont}$ is the mean-squared error between the regressed values and the continuous emotions. Finally $\mathcal{L}_{emb}$ is as in (\ref{eq:embedloss}).

\section{Experimental Results}
We train each TSN for 50 epochs using Stochastic Gradient Descent (SGD), with initial learning rate $10^{-3}$ which drops by a factor of 10 at $20$ epochs\footnote{PyTorch code available at \href{https://github.com/filby89/NTUA-BEEU-eccv2020}{https://github.com/filby89/NTUA-BEEU-eccv2020}
}. The backbone networks used is a residual network (ResNet) with 101 layers for the body convNets and a ResNet with 50 layers for the context convNet. We use the default hyperparameters of TSNs: 3 segments, 1 frame from each segment for the RGB streams, and 5 frames from each segment for the optical flow stream. The consensus used for segment fusion is averaging. For each network, we select the epoch with the best validation ERS. We have also found experimentally that the partialBN (Batch Normalization) technique used in \cite{wang2016temporal} gives a nontrivial boost to the performance of the network.

First, in Table~\ref{tab:ablation} we present two ablation experiments regarding the addition of $\mathcal{L}_{emb}$. We can see that adding the embedding loss increases slightly the performance in the RGB-b stream, and gives a boost to the performance of the Flow-b stream.



\begin{table}[t]
\centering
\setlength\tabcolsep{8pt}
\def\arraystretch{1.5}
\begin{tabular}{l|l|c|c|c|c}
 & Model & \textbf{$mAP$}    & \textbf{$mRA$}    & \textbf{$mR^2$} & \textbf{$ERS$}    \\ \hline \hline 

\multirow{3}{*}{without $\mathcal{L}_{emb}$} & RGB-b                  & 0.1567          & 0.6140          & 0.0538                         & 0.21955      \\
& Flow-b                 & 0.1444          & 0.5914          & 0.0507                         & 0.2093          \\ 
& RGB-b + Flow-b                    & 0.1623          & 0.6307          & 0.078                          & 0.2375          \\ \hline

\multirow{3}{*}{with $\mathcal{L}_{emb}$} &RGB-b                  & 0.1564          & 0.6143          & 0.0546                         & 0.21997           \\ 
&Flow-b                 & 0.1465          & 0.5947          & 0.0579                         & 0.2142          \\ 
& RGB-b + Flow-b                   & \textbf{0.1637} & \textbf{0.6327} & \textbf{0.0874}                & \textbf{0.2428} \\ \hline

\end{tabular}
\caption{Ablation experiment by training with and without $\mathcal{L}_{emb}$.}
\label{tab:ablation}
\end{table}

Then, in Table \ref{tab:ab} we present our experimental results on the validation set of BoLD including the RGB context stream. From the results we can see that including the context along with the body in the RGB modality boosts the validation ERS of the architecture. We also experimented with including the context in the Flow network, but this resulted in worse performance. Our final submission for the test set was the model with the best validation score (0.2439 employing RGB-bc + Flow-b), using 25 segments instead of 3. The results of the different metrics on the test set can also be seen in Table \ref{tab:ab}, while the final ERS is 0.26235, improving upon the previous best result of 0.2530\cite{luo_arbee_2020}.

\begin{table}[t]
\centering
\setlength\tabcolsep{8pt}
\def\arraystretch{1.5}
\begin{tabular}{l|l|c|c|c|c}
set& Model & \textbf{$mAP$}    & \textbf{$mRA$}    & \textbf{$mR^2$} & \textbf{$ERS$}    \\ \hline \hline 

\multirow{3}{*}{valid} & RGB-c            & 0.1395    & 0.5760    & 0.0365                      & 0.1971     \\
& RGB-bc           & 0.1566       & 0.6055       & 0.0675                         & 0.2243       \\
& RGB-bc + Flow-b  & 0.1656    & 0.6266    & 0.0917                      & \textbf{0.2439 }  \\ \hline\hline
test & RGB-bc + Flow-b & 0.1796 & 0.6416 & 0.1141 & \textbf{0.26235} \\
\hline
\end{tabular}
\caption{Results on the validation and test set of BoLD including the RGB context stream and $\mathcal{L}_{emb}$.}
\label{tab:ab}
\end{table}

\section{Conclusions}
In this paper we presented our method submitted at the BEEU challenge, winning first place. Our method extended the TSN framework to include a visual-semantic embedding loss, by utilizing GloVE word embeddings, and also included an additional context stream for the RGB modality. We verified the superiority of our extensions compared to the baseline on the validation set of the challenge, and submitted the best system which achieved 0.26235 Emotion Recognition Score on the BoLD test set, surpassing the previous best result of 0.2530.

\section*{Acknowledgments}
This research is carried out / funded in the context of the project ``Intelligent Child-Robot Interaction System for designing and implementing edutainment scenarios with emphasis on visual information" (MIS  5049533) under the call for proposals ``Researchers' support with an emphasis on young researchers- 2nd Cycle”. The project is co-financed by Greece and the European Union (European Social Fund- ESF) by the Operational Programme Human Resources Development, Education and Lifelong Learning 2014-2020.

%
%
\bibliographystyle{splncs04}
\bibliography{egbib}
\end{document}